\documentclass[lettersize,journal]{IEEEtran}
\usepackage{amsmath,amsfonts}
\usepackage{algorithmic}
\usepackage{algorithm}
\usepackage{array}
\usepackage[caption=false,font=normalsize,labelfont=sf,textfont=sf]{subfig}
\usepackage{textcomp}
\usepackage{stfloats}
\usepackage{url}
\usepackage{verbatim}
\usepackage{listings}
\usepackage{graphicx}
\usepackage{cite}
\usepackage{todonotes}
\usepackage{xcolor}
\usepackage{amsmath}
\usepackage{amssymb}
\usepackage{multirow}

\usepackage{colortbl}
\newcommand{\col}{\cellcolor[HTML]{EFEFEF}}
\newcommand{\mstd}[2]{\ensuremath{#1{\scriptstyle \pm #2}}}
\newcommand{\mstdb}[2]{\ensuremath{\textbf{#1}{\scriptstyle \pm \textbf{#2}}}}

\begin{document}

\title{Analyzing the Behaviour of Tree-Based Neural Networks in Regression Tasks}

\author{
    \IEEEauthorblockN{Peter Samoaa\IEEEauthorrefmark{1},
    Mehrdad Farahani\IEEEauthorrefmark{1}, Antonio Longa\IEEEauthorrefmark{3}, Philipp Leitner\IEEEauthorrefmark{2}, 
    Morteza Haghir Chehreghani\IEEEauthorrefmark{1}}
    
    \IEEEauthorblockA{\IEEEauthorrefmark{1}Data Science and AI Division\\
    Chalmers University of Technology\\
    Gothenburg, Sweden\\
    Email: \{samoaa, mehrdad.farahani, morteza.chehreghani\}@chalmers.se}
    
    \IEEEauthorblockA{\IEEEauthorrefmark{2}Interaction Design and Software Engineering Division\\
    Chalmers University of Technology\\
    Gothenburg, Sweden\\
    Email: philipp.leitner@chalmers.se}
    
    \IEEEauthorblockA{\IEEEauthorrefmark{3}Department of Information Engineering and Computer Science\\
    University of Trento\\
    Trento, Italy\\
    Email: antonio.longa@unitn.it}
}

\markboth{This work has been submitted to the IEEE for possible publication.}
{Copyright may be transferred without notice, after which this version may no longer be accessible.}


\maketitle

\begin{abstract}


The landscape of deep learning has vastly expanded the frontiers of source code analysis, particularly through the utilization of structural representations such as Abstract Syntax Trees (ASTs). While these methodologies have demonstrated effectiveness in classification tasks, their efficacy in regression applications, such as execution time prediction from source code, remains underexplored. This paper endeavours to decode the behaviour of tree-based neural network models in the context of such regression challenges. We extend the application of established models—tree-based Convolutional Neural Networks (CNNs), Code2Vec, and Transformer-based methods—to predict the execution time of source code by parsing it to an AST. Our comparative analysis reveals that while these models are benchmarks in code representation, they exhibit limitations when tasked with regression. To address these deficiencies, we propose a novel dual-transformer approach that operates on both source code tokens and AST representations, employing cross-attention mechanisms to enhance interpretability between the two domains. Furthermore, we explore the adaptation of Graph Neural Networks (GNNs) to this tree-based problem, theorizing the inherent compatibility due to the graphical nature of ASTs. Empirical evaluations on real-world datasets showcase that our dual-transformer model outperforms all other tree-based neural networks and the GNN-based models. Moreover, our proposed dual transformer demonstrates remarkable adaptability and robust performance across diverse datasets.

\end{abstract}


\section{Introduction}

Deep learning models have widely used in the source code analysis for various tasks such as classification of source code~\cite{Bui2018,kanade2020,CNN_tree}, detection of code clones~\cite{Bui:21-2,Fang2020,Mehrotra2020}, identification of bugs~\cite{Hua2021,Li2021ICSE,Shi2021}, and generation of code summaries~\cite{Bui:21-1,Zhang:20,liu2021}. 

Source code can be represented as textual format, thereby encapsulating the lexical content of the code~\cite{samoaaSLR2022}. Through the textual representation of the source code we can extract the lexical information. For that aim, most traditional approaches to processing source code often adopt string-based pattern matching, rule-based model transformation, and bag-of-words~\cite{ALSABBAGH2022}. However, these methods treat code fragments as plain texts, which ignore the underlying semantic information in source code, resulting in poor performance.

Source code can be represented as a tree throughout the abstract syntax tree (AST)~\cite{samoaaSLR2022}. Thus, the code can be parsed to the tree directly without prior execution. AST representation is abstract and does not include all available details, such as punctuation and delimiters. Theoretically, ASTs can be used to illustrate the syntactic structure of source code, such as the function name and the flow of the instructions (for example, in an if or while construct). 

Source code can also be represented as a graph which explains the semantic information from the source code~\cite{samoaaSLR2022}. The graph-structured representations can only be extracted via the intermediate representation or bytecode~\cite{Zhao2018} (e.g. control flow graphs which describe the sequence in which the instructions of a program will be executed~\cite{LIC2021}, data flow graphs which follow and tracks the usage of the variables through the program~\cite{LIC2021}, call flow graphs which captures the relation between a statement which calls a function and the called function~\cite{cummins2020programl}), which means that the code fragments have to be compiled successfully. However, the graphs may contribute to enriching code representations. Unfortunately, arbitrary code fragments or incomplete code snippets usually lose the import libraries or dependency packages, making the compilation fail. Such a limitation may make a large number of labelled code snippets unavailable for training, hindering the application of graph representations in practice~\cite{hua2022transformer}.

\textit{That is why, in our study, we will focus on trees as they are easier to extract since we just need to parse the source code. }

Recently, some approaches combined neural networks and ASTs to constitute tree-based neural networks  (TNNs)~\cite{AST_NN}. Given a tree, TNNs learn the vector representation by recursively computing node embeddings in a bottom-up way. Popular TNN models are the Recursive Neural Network (RvNN)~\cite{DLCF}, Tree-based convolutional neural networks  (TBCNN)~\cite{CNN_tree}, and Tree-based Long Short-Term Memory (Tree-LSTM)~\cite{SDF}. However, most TBNN approaches tackle a classification problem for the source code but not regression tasks. Regression tasks such as source code performance prediction (predicting the execution time for the application prior to running it) can give the developer an early indication of the runtime behaviour of their source code. Samoaa et al.~\cite{SamoaaTEPGNN} indicate that trying the TNNs approaches for regression tasks will lead to poor efficiency compared to classification tasks. That said, these solutions are not generic enough for any source code analysis tasks. Thus, to understand the behaviour of TBNN models in regression tasks, we design an analytical framework that uses the benchmark TBNN models for source code analysis to prove the claim that these models are efficient in classification tasks but in a regression context. The TBNN models that are used in our framework are code2vec~\cite{Alon19}, TBCNN~\cite{CNN_tree}, and Transformer-based networks over AST(TBAST)~\cite{hua2022transformer}, taking into account that we have to make some changes in the architecture of these models to fit the regression tasks. 
Additionally, we explore various GNN architectures, focusing on neighbourhood information sampling and aggregation within the AST, to further enrich our analysis framework.
 
To address the lack of efficiency of these TBNN models in a regression context, we develop our model based on cross-attention dual transformers, which utilize sequences of source code tokens and AST nodes. By employing cross-attention mechanisms between the two transformers, our model aims to elucidate the influence of individual source code tokens on AST nodes, enhancing the understanding of code semantics. 

Then, we analyse the behaviour of each type of architecture (convolution, sequence, and GNN) for different levels of information: node level (for every node in the AST) as in TBCNN, GNNs, and sequential transformers or path level (a path in AST, which is a sequence of nodes) as in code2vec. 
Since we have a regression value for each source code program, we have to map each AST to the regression value. Thus, the AST has to be represented as one vector. For that aim, we will aggregate the node and path representations through the models into one continuous vector. 
To increase the reliability, we use two different real-world datasets of performance measurements. The first dataset (\emph{OSSBuild}) is real build data collected from the continuous integration systems of four open-source systems. The second (\emph{HadoopTests}) is a larger dataset which we have collected ourselves by repeatedly executing unit tests from the Hadoop open-source system in a controlled environment. 

The key findings of our experiments show that our dual-transformers model consistently outperformed traditional TBNN and GNN models across various metrics, including Mean Squared Error (MSE), Mean Absolute Error (MAE), and Pearson correlation. This superiority was observed in both dataset contexts (OssBuilds and Hadoop) and under different experimental setups, such as varying training sizes and cross-dataset transferability.
In addition, The dual-transformers model demonstrated remarkable adaptability and robust performance across diverse datasets. This model effectively handled the complexity and variability of datasets differing in size and composition, indicating their potential for general application not only in source code analysis but also in other tree data domains. The study also underscored how the characteristics of datasets, such as the diversity of the data and the structure of ASTs, significantly affect the performance of neural network models. This was evident from the varying performances of models on the OssBuilds dataset, which comprises data from multiple projects, compared to the Hadoop dataset, which is more homogeneous.

The aforementioned key findings highlight the potential of our study in understanding the behaviour of different types of neural network architectures for regression tasks. Our contributions are manifold and address several gaps in the current landscape of tree-based neural network methodologies for regressions:
\begin{enumerate}
    \item \textbf{Novel Transformer-Based Model for Tree Learning:} We addressed the inefficiency of different used models for tree and regression by designing and implementing a model-centric AI for employing both code tokens and tree nodes in the transformer based on cross-attention.
    \item \textbf{Development of Specialized Tree Datasets:} We propose new tree datasets designed to be directly usable by researchers, facilitating further exploration and validation of tree-based models.
    \item \textbf{Novel Framework for Analyzing the Behaviour of Different Tree-Based Neural Networks Models :} We provide the research community with an open-source framework that merges all TBNN models with our dual-transformers model. So, researchers can directly use this framework for different tasks and research. The code files are available on GitHub~\url{https://github.com/petersamoaa/Tree_based_NN_Error_analysis}, and the data files are available on Zenodo~\cite{samoaa_2024_11383081} 
\end{enumerate}

\section{Background}
\subsection{Abstract syntax trees}
Abstract Syntax Trees (ASTs) offer a hierarchical representation of source code that abstracts away from the specific syntactic form, focusing instead on the underlying structure and logic of the code. This representation discards superficial elements like punctuation, concentrating on the nodes that signify the fundamental constructs of programming languages, such as variables, operators, and control structures.

An AST encapsulates the syntactic structure of code, where each node represents a construct occurring within the source code~\cite{samoaa2023_bigdata}. The tree's edges outline the relationship between these constructs, effectively mapping out the syntax rules of the language. The root of the tree often represents the entire program, with leaves corresponding to atomic elements like literals or variable names and internal nodes representing operator or control statements that dictate the flow of the program~\cite{samoaaSLR2022}.

Transforming source code into an AST involves parsing, where the code is analyzed according to the grammar of the programming language, and its structure is broken down into a tree that reflects the hierarchical composition of the code's elements. This process facilitates various code analysis tasks by providing a structured and navigable representation of the code, enabling more sophisticated and accurate analyses than linear source code examination.

ASTs are instrumental in various applications, from code compilation and optimization to more advanced analyses like static code analysis, refactoring, and understanding program behaviour. By providing a structured view of code, ASTs allow tools and developers to examine the abstract properties of the program without getting bogged down by syntactic details irrelevant to the analysis at hand.

In the context of programming language analysis, especially with the advent of machine learning techniques, ASTs serve as a crucial bridge between source code and its semantic understanding. They enable the application of advanced analytical models that can learn from the structural patterns of code to perform tasks such as bug detection, code summarization, and even automated code generation~\cite{samoaaSLR2022}.

\subsection{Motivation Example}
To have a deeper understanding of the AST, this section explains by example how the source code can be represented as AST. Thus, we investigate Java source code files (see Listing~\ref{java:example}). 

\begin{lstlisting}[float=h!, language=Java, caption=A Simple JUnit 5 Test Case, label=java:example, basicstyle=\scriptsize]
package org.myorg.weather.tests;

import static
    org.junit.jupiter.api.Assertions.assertEquals;
import org.myorg.weather.WeatherAPI;
import org.myorg.weather.Flags;

public class WeatherAPITest {
    
    WeatherAPI api = new WeatherAPI();
    
    @Test
    public void testTemperatureOutput() {
        double currentTemp = api.currentTemp();
        Flags f = api.getFreezeFlag();
        if(currentTemp <= 3.0d)
            assertEquals(Flags.FREEZE, f);
        else
            assertEquals(Flags.THAW, f);
    }
}
\end{lstlisting}

In this example, a single test case \texttt{testTemperature\-Output()} is presented that tests a feature of an (imaginary) API. As common for test cases, the example is short and structurally relatively simple. Much of the body of the test case consists of invocations to the system-under-test and calls of JUnit standard methods, such as \texttt{assertEquals}.

\begin{figure}
    \centering
    \includegraphics[width=\linewidth]{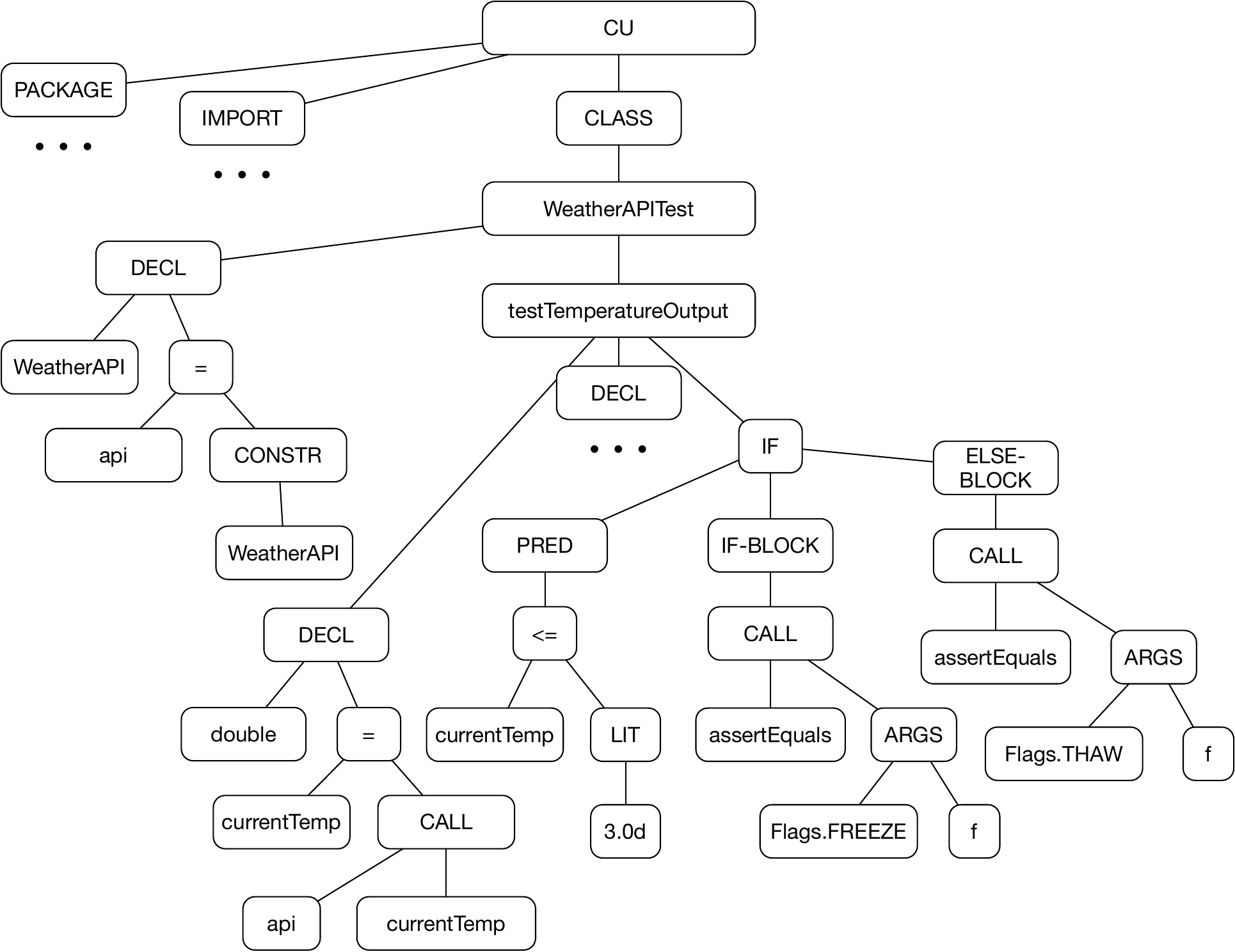}
    \caption{Simplified abstract syntax tree (AST) representing the illustrative example presented in Listing~\ref{java:example}. Package declarations, import statements, as well as the declaration in Line 15 are skipped for brevity.}
    \label{fig:AST_draft}
\end{figure}

A (slightly simplified) AST for this illustrative example is depicted in Figure~\ref{fig:AST_draft}. The produced AST does not contain purely syntactical elements, such as comments, brackets, or code location information. We make use of the pure Python Java parser javalang\footnote{https://pypi.org/project/javalang/} to parse each test file and use the node types, values, and production rules in javalang to describe our ASTs.

\section{Related Work}
The application of deep learning techniques to tree-structured representations of source code has garnered considerable attention within the research community. Mou et al.\cite{CNN_tree,Mou_Li_Zhang_Wang_Jin_2016} introduced a novel approach leveraging tree-based Convolutional Neural Networks (CNNs) to perform convolutional computations on Abstract Syntax Trees (ASTs) for code classification tasks. Similarly, Zhu et al.\cite{zhub15} employed tree-based Long Short-Term Memory (LSTM) networks to encode AST pairs into continuous vectors, facilitating code clone detection by measuring similarities.

Further exploring tree-based neural networks, Zhang et al.\cite{AST_NN} utilized Recursive Neural Networks (RvNNs) to process ASTs at the path level, targeting classification objectives. Concurrently, While et al.\cite{DLCF} applied RvNNs to analyze ASTs at the node level for classification purposes, a method paralleled by Wei et al.~\cite{SDF} through the use of Labeled AST (LAST) structures.

Innovatively, Zhang et al.~\cite{ZHANG2023} introduced a transformer-based model that incorporates tree-based position embeddings to represent the nodes within ASTs, enhancing the classification of source code by learning from code tokens.

Beyond classification, the generation of source code has also been explored. A novel pre-trained model, TreeBERT~\cite{treeBERT2021}, adapts the BERT architecture to understand programming languages through AST analysis, focusing on path-level node position embeddings for code summarization tasks. Yang et al.~\cite{MMTrans2021} further this exploration by proposing the use of multi-modal transformers, analyzing ASTs at the node level for code summarization.

In the realm of regression tasks, the work of Samoaa et al.~\cite{SamoaaTEPGNN} stands out by applying Graph Neural Networks (GNNs) to augmented ASTs, representing a pioneering effort in leveraging tree-based neural network models for regression in source code analysis.

Despite the proliferation of deep learning methodologies for analyzing source code through AST representations, there remains a gap in the literature concerning the comparative analysis of different architectural approaches, particularly in the context of regression tasks. This study aims to bridge this gap by examining the behaviour and efficacy of various tree-based neural network models in regression scenarios.

\section{Analytic Framework}

\begin{figure}[!h]
    \centering
    \includegraphics[width=\linewidth]{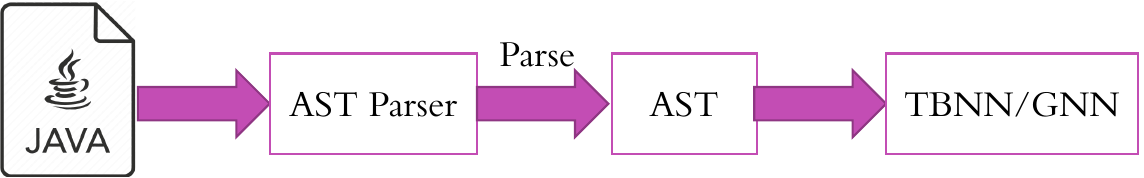}
    \caption{Abstracted General Code Representation and DL Models in Software Engineering.}
    \label{fig:TBNN}
\end{figure}

According to Samoaa~\cite{samoaaSLR2022}, the majority of the deep-learning-based approaches for source code follow the same pipeline as in Figure~\ref{fig:TBNN}. Thus, the approaches start with parsing the code into AST through the AST parser. Our study uses a Python Java parser javalang\footnote{https://pypi.org/project/javalang/} as a parser that produces AST from the source code. AST represent the syntactic features. Then, deep learning models like recurrent neural networks (RNNs) or convolutional neural networks (CNNs) are used to encode the nodes of AST into vectors for downstream tasks like classification and regression.

These approaches have three major limitations: 1) RNN models inevitably suffer from the gradient vanishing problem, meaning that the gradients become vanishingly small values during model training, especially in the context of usage of AST which is very deep in most cases~\cite{mou2018tree}; 2) CNN models cannot capture the long-distance dependency information from sequential nodes of tokens in AST due to the size limitation of the sliding windows, which scan only a few nodes/tokens at a time~\cite{mou2018tree}; 3) apart from using the simplistic lexical features, the approaches for AST processing that recursively traverse entire trees from bottom to top may produce longer sequential inputs than the textual inputs, consume large amounts of computational resources and destroy the syntactical structures existing in AST~\cite{hua2022transformer}

Thus, based on the abovementioned limitations, we will use a transformer as a competitor for sequential models as well as the basis of our approach since also the transformer is the most mature of sequential models for the following reasons:

\begin{itemize}
    \item \textbf{Handling Long Dependencies:} Transformers leverage self-attention mechanisms. This allows them to weigh the importance of different parts of the input sequence directly, regardless of the distance between elements~\cite{attention}, making them well-suited for the hierarchical and complex structures of ASTs.
    \item \textbf{Parallelization:} Transformers do not process data sequentially as RNNs do. Instead, they can process entire data sequences in parallel during training, significantly speeding up the learning process. This is particularly advantageous when dealing with the large and intricate structures of ASTs, where computational efficiency is paramount.
    \item \textbf{Flexibility in Capturing Structural Information: }The self-attention mechanism in transformers can easily adapt to the structured nature of ASTs, capturing both the local and global context within the tree. This flexibility allows for a more nuanced understanding of code semantics compared to the fixed window size of CNNs or the sequential nature of RNNs.
    \item \textbf{Scalability:} Transformers are highly scalable and capable of handling large input sequences without significantly dropping performance. This makes them ideal for source code analysis, where ASTs can vary widely in size and complexity.
\end{itemize}

Despite its mentioned limitation, we will also use CNN in our framework to have diverse architecture types of neural networks. 

\section{Dual Transformer Model}
\label{sec:dualtransformer}
Most models use attention except the TBCNN. The main novelty of our work through the designing and developing of our approach, as well as the comparison with benchmark models, is the understanding that attention mechanism over multiple contexts is needed for embedding programs into a continuous space, and the use of this embedding to predict properties of a whole code snippet.

\begin{figure}
    \centering
    \includegraphics[width=0.5\textwidth]{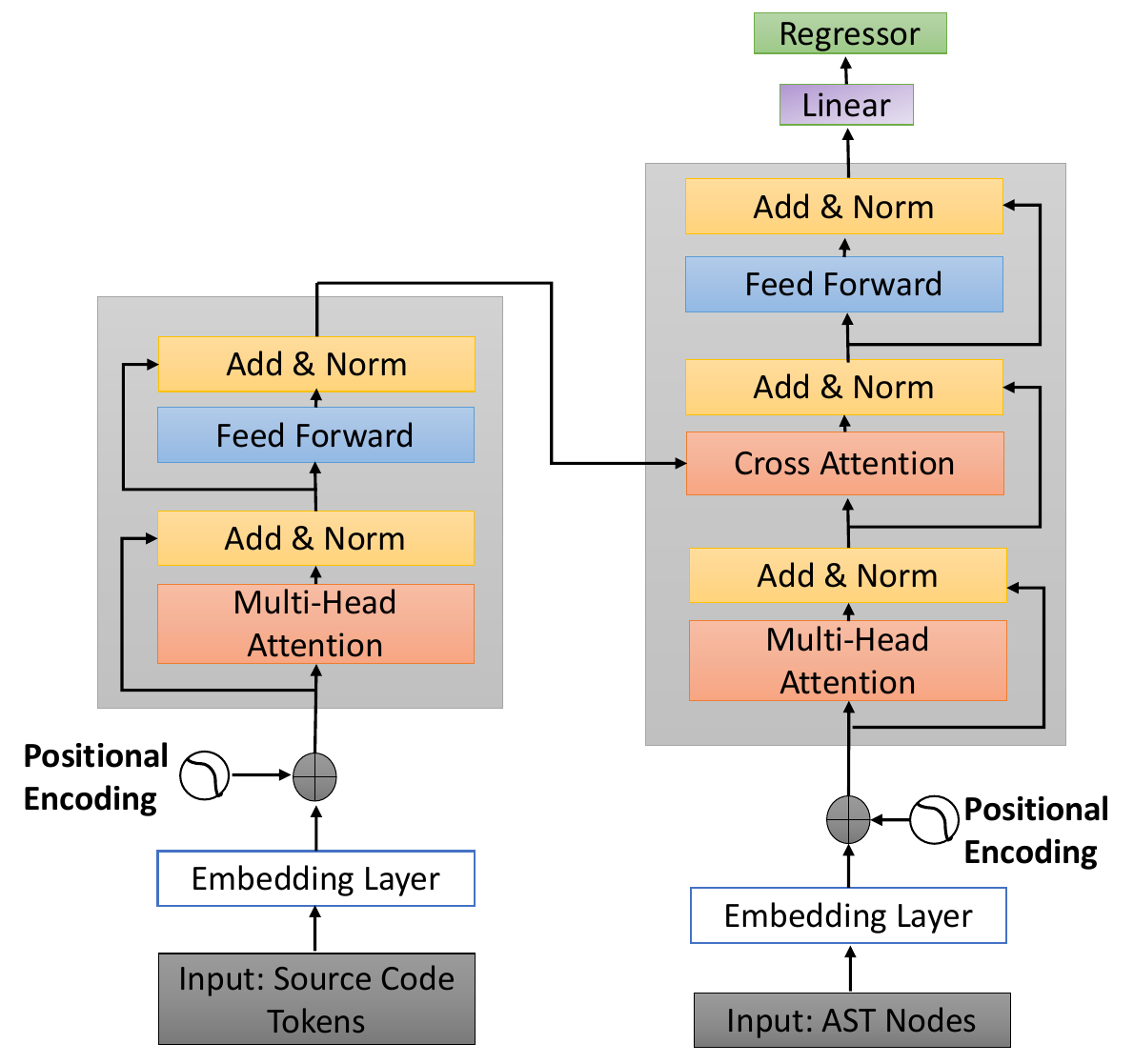}
    \caption{The architecture of the Dual-Transformer model. The framework features two transformer encoders: NLEncoder for source code tokens and ASTEncoder for AST nodes, each with layers for embedding, multi-head attention, and feed-forward networks, complemented by add \& norm layers for stabilization. Their outputs are merged via cross-attention and passed to a linear regressor for error prediction, leveraging both textual and syntactical insights.}
    \label{fig:dual_transformer_arch}
\end{figure}

Our Dual-Transformer model is designed to integrate structural and lexical information within the source code to predict execution time. As illustrated in Figure \ref{fig:dual_transformer_arch}, the architecture consists of two parallel transformer encoders: the NLEncoder for processing source code tokens and the ASTEncoder for processing AST node that the outputs of both encoders are integrated via a cross-attention mechanism, which allows the model to jointly consider textual and structural information. 

\subsection{NL-Encoder}
The NL-Encoder serves to encode textual information from source code tokens. Input tokens $x_{code}$ are transformed into embeddings $E_{code}$ via a learned embedding matrix $W_{code}$, combined with positional encodings $P_{code}$ to retain sequential information:
\begin{equation}
    \label{eq:001}
    E_{code} = W_{code} \cdot x_{code} + P_{code}
\end{equation}

These embeddings then pass through a series of transformer blocks, each comprising a multi-head self-attention mechanism and a position-wise feed-forward network. For the $i^{th}$ block, the output $O_i$ is computed as follows:

\begin{align}
    \label{eq:002}
    O'_{i} &= \text{LayerNorm}(E_{code} + \text{MultiHead}(E_{code}, E_{code}, E_{code})) \\
    O_{i} &= \text{LayerNorm}(O'_{i} + \text{FFN}(O'_{i}))
\end{align}

Where LayerNorm denotes layer normalization, MultiHead denotes the multi-head self-attention mechanism and FFN represents the feed-forward network. The embeddings are subsequently refined by transformer layers, with the output of the final layer being denoted as $O_{code}$.




\subsection{AST-Encoder}
The ASTEncoder parallels the NLEncoder in structure but operates on the AST's nodes. Similar to the NLEncoder, AST node inputs $x_{ast}$ are embedded into vectors $E_{ast}$ and supplemented with positional encoding:

\begin{equation}
    \label{eq:004}
    E_{ast} = W_{ast} \cdot x_{ast} + P_{ast}
\end{equation}

These embeddings are then processed through analogous transformer blocks, yielding a structured representation of the code's syntax as $O_{ast}$.



\subsection{Attention Mechanisms}

The crux of our model lies in the cross-attention mechanism that bridges the NLEncoder and ASTEncoder. For each pair of encoded sequences $O_{code}$ and $O_{ast}$, cross-attention is computed as:


\begin{align}
    \text{CrossAttention}&(O_{\text{code}}, O_{\text{ast}}) \notag \\
    &= \text{Attention}(O_{\text{code}}W^Q_{\text{cross}}, O_{\text{ast}}W^K_{\text{cross}}, O_{\text{ast}}W^V_{\text{cross}}) \notag \\
    &= \text{softmax}\left(\frac{(O_{\text{code}}W^Q_{\text{cross}})(O_{\text{ast}}W^K_{\text{cross}})^T}{\sqrt{d_k}}\right)O_{\text{ast}}W^V_{\text{cross}} \label{eq:cross_attention}
\end{align}

Where the learned weight matrices $W^Q_{cross}$, $W^K_{cross}$, and $W^V_{cross}$ are central to the model's ability to integrate the outputs of the NLEncoder and ASTEncoder. These matrices transform the final layer outputs of the encoders into the queries (Q), keys (K), and values (V) needed for the attention calculation. This allows each encoder to attend to the outputs of the other, integrating semantic and syntactic information into a unified representation. \\




\subsection{Regression Head}
At the top of the model, a regression head is applied to integrate the representation ($z$) of the output of ASTEncoder for error prediction:
\begin{equation}
     \hat{y} = \text{Linear}(\text{ReLU}(\text{Linear}(z)))
\end{equation}

Where $z$ represents the result produced by the first token "[CLS]" from the ASTEncoder, which is designed to summarize the overall context of the input sequence.

\section{Other GNN and TBNN Models}
\label{sec:othermodels}
This section introduces the benchmark models against which our dual transformers model is evaluated. This includes discussing GNN-based models, a convolutional model leveraging tree structures, a sequential model transformer-based, and the path-attention mechanism employed by code2vec. Each approach offers a unique perspective on source code analysis through AST, setting the stage for a comprehensive comparative study.

\subsection{Graph Learning Approach}
Graph Neural Networks have demonstrated promise in various real-world applications\cite{buffelli2021graph,bianchi2024expressive,pasa2023unified,ferrini2023metapath,nguyen2022emotion,telyatnikov2023hypergraph,thomas2023graph,tiezzi2022graph}. Two primary models have played a pioneering role in the field, establishing the foundational frameworks for two key approaches to graph processing: the recurrent model proposed by Scarselli et al.\cite{scarselli2008graph} and the feedforward model introduced by Micheli\cite{micheli2009neural}. Notably, the feedforward approach has evolved into the prevailing method\cite{kipf2017semisupervised,GATvelickovic2018graph,bacciu2020gentle,SAGEhamilton2018inductive,GINxu2019powerful}.

In this section, we will explain the graph neural network architectures that we used in our experiment. The models accept the AST as an input and predict a scalar execution time value.
\paragraph{GCN (Graph Convolutional Network)}

GCNs~\cite{kipf2017semisupervised} leverage the concept of convolutional operations on graph-structured data. The model updates the representation of a node by aggregating the features of its neighbours.

\begin{equation}
H^{(l+1)} = \sigma(\tilde{D}^{-\frac{1}{2}} \tilde{A} \tilde{D}^{-\frac{1}{2}} H^{(l)} W^{(l)})
\end{equation}

Where $H^{(l)}$ is the matrix of node features at layer $l$, $\tilde{A} = A + I_N$ is the adjacency matrix $A$ with added self-connections $I_N$, $\tilde{D}$ is the degree matrix of $\tilde{A}$, $W^{(l)}$ is the weight matrix for layer $l$, and $\sigma$ is a non-linear activation function.

\paragraph{GAT (Graph Attention Network)}
GAT~\cite{GATvelickovic2018graph} introduces the attention mechanism to graph neural networks. It computes the hidden representations of each node by attending to its neighbours, following a self-attention strategy.

\begin{equation}
\alpha_{ij} = \frac{\exp(\text{LeakyReLU}(a^T [W h_i || W h_j]))}{\sum_{k \in \mathcal{N}(i)} \exp(\text{LeakyReLU}(a^T [W h_i || W h_k]))}
\end{equation}
\begin{equation}
h'_i = \sigma\left(\sum_{j \in \mathcal{N}(i)} \alpha_{ij} W h_j\right)
\end{equation}

Where $h_i$ is the feature vector of node $i$, $W$ is a shared linear transformation, $a$ is the attention mechanism's learnable weight, $||$ denotes concatenation, and $\alpha_{ij}$ represents the attention coefficient between nodes $i$ and $j$.

\paragraph{GraphSAGE (Graph Sample and Aggregation)}
GraphSAGE~\cite{SAGEhamilton2018inductive} generates embeddings by sampling and aggregating features from a node's local neighbourhood.

\begin{equation}
h'_i = \sigma\left(W \cdot \text{MEAN}(\{h_i\} \cup \{h_j, \forall j \in \mathcal{N}(i)\})\right)
\end{equation}

Where $h_i$ is the feature vector of node $i$, $\mathcal{N}(i)$ is the set of its neighbours, and $W$ is the weight matrix associated with the aggregator function.

\paragraph{GIN (Graph Isomorphism Network)}

GIN~\cite{GINxu2019powerful} is designed to capture the power of the Weisfeiler-Lehman graph isomorphism test. It aggregates neighbour information to update node representations, aiming to distinguish graph structures.

\begin{equation}
h'_i = \text{MLP}\left((1 + \epsilon) \cdot h_i + \sum_{j \in \mathcal{N}(i)} h_j\right)
\end{equation}

Where $h_i$ represents the feature vector of node $i$, $\epsilon$ is a learnable parameter or a fixed scalar, and MLP denotes a multi-layer perceptron.

Since all baselines are used for classification, we changed the models to fit the regression tasks. 
\subsection{Tree-based CNN (TBCNN)}
TBCNN models~\cite{CNN_tree} are designed to process the structured data of an AST by leveraging convolutional layers tailored for tree structures. This approach involves several key components:
\begin{itemize}
    \item \textbf{Representation Learning for AST Nodes:} Each AST node is represented as a distributed vector capturing the symbol features. 
    \begin{equation}
    \vec{p} \approx \tanh\left(\sum_{i} l_i \mathbf{W}_{\text{code}, i} \cdot \vec{c_i} + \mathbf{b}_{\text{code}}\right) \tag{1}
\end{equation}
Where:

$\vec{p}$ is the parent node's vector representation.
$l_i$ is a coefficient based on the subtree's leaf count.
$\mathbf{W}{\text{code}, i}$ is learned weight matrices.
$\vec{c_i}$ and $\vec{x_i}$ represent the children nodes' vectors.
\item \textbf{Coding Layer:} This layer encodes the representation of a node by aggregating the features of its children through a learned transformation.

\begin{equation}
\vec{p} = \mathbf{W}_{\text{comb1}} \cdot \vec{p} + \mathbf{W}_{\text{comb2}} \cdot \tanh\left(\sum_{i} l_i \mathbf{W}_{\text{code}, i} \cdot \vec{c_i} + \mathbf{b}_{\text{code}}\right) \tag{2}
\end{equation}
where:
 $\mathbf{W}{\text{comb1}}$ and $\mathbf{W}{\text{comb2}}$ are learned weight matrices.
\item \textbf{Tree-based Convolutional Layer:} A set of convolutional filters or kernels is applied over the AST to capture the hierarchical structure of the code.
\begin{equation}
\mathbf{y} = \tanh\left(\sum_{i} \mathbf{W}_{\text{conv}, i} \cdot \vec{x_i} + \mathbf{b}_{\text{conv}}\right) \tag{3}
\end{equation}
where  $\mathbf{W}{\text{conv}, i}$ is learned weight matrices.
$\mathbf{b}{\text{code}}$ and $\mathbf{b}{\text{conv}}$ are bias terms.
$\mathbf{y}$ is the output vector after applying the convolution operation.
\item \textbf{Dynamic Pooling:} This layer aggregates the convolutional features from different parts of the AST to handle varying sizes and shapes.

\item \textbf{The "Continuous Binary Tree" Model:} It addresses the challenge of AST nodes having varying numbers of children by considering each subtree as a binary tree during convolution.
\end{itemize}

This representation captures the essence of how TBCNNs operate on ASTs to learn meaningful representations of source code for various tasks such as program classification and pattern detection. However, we modified the model's architecture to fit regression tasks.

\subsection{code2vec Path-Attention Model}




The code2vec model operates on the principle of transforming code snippets into a distributed vector representation. It achieves this by embedding the paths and terminal nodes of AST and using an attention mechanism to identify and aggregate the most relevant features. The model can be decomposed into several components:

\begin{itemize}
    \item \textbf{Embedding Vocabularies}: Two embedding matrices, $value\_vocab \in \mathbb{R}^{|X| \times d}$ and $path\_vocab \in \mathbb{R}^{|P| \times d}$, where $|X|$ is the number of unique AST terminal node values and $|P|$ is the number of unique AST paths observed during training. The embedding size $d$ is a hyperparameter typically ranging between 100 and 500.
    
    \item \textbf{Context Vectors}: A path-context $b_i$ is a triplet $\langle x_s, p_j, x_t \rangle$ representing the start and end tokens of a path in the AST and the path itself. Each component of $b_i$ is mapped to its embedding and concatenated to form a context vector $c_i \in \mathbb{R}^{3d}$:
    
    \begin{multline}
c_i = \text{embedding} \langle x_s, p_j, x_t \rangle \\
= \left[ value\_vocab[s], path\_vocab[j], value\_vocab[t] \right] \in \mathbb{R}^{3d}
\end{multline}

    \item \textbf{Fully Connected Layer}: Each context vector $c_i$ is transformed by a fully connected layer with weights $W \in \mathbb{R}^{d \times 3d}$ and a $\tanh$ non-linearity to produce a combined context vector $\tilde{c}_i$:
    
    \begin{equation}
        \tilde{c}_i = \tanh(W \cdot c_i)
    \end{equation}
    
    \item \textbf{Attention Mechanism}: The attention mechanism computes a weighted average of the combined context vectors $\tilde{c}_i$, using an attention vector $a \in \mathbb{R}^{d}$ which is learned during training. The attention weight $\alpha_i$ for each $\tilde{c}_i$ is computed using the softmax function:
    
    \begin{equation}
        \alpha_i = \frac{\exp(\tilde{c}_i^\top \cdot a)}{\sum_{j=1}^{n} \exp(\tilde{c}_j^\top \cdot a)}
    \end{equation}
    
    \item \textbf{Aggregated Code Vector}: The final code vector $v \in \mathbb{R}^{d}$ representing the entire code snippet is calculated as a weighted sum of the combined context vectors:
    
    \begin{equation}
        v = \sum_{i=1}^{n} \alpha_i \cdot \tilde{c}_i
    \end{equation}
    
\end{itemize}

The model learns to assign an appropriate amount of attention to each path context, effectively capturing the semantics of the code snippet. The final code vector can be used for various downstream tasks, such as method name prediction, with the attention weights offering insight into the model's decision process.

\subsection{Transformer-based Networks for AST}

This approach splits the deep ASTs into smaller subtrees that aim to exploit syntactical information in code statements. Then, the model gets the sequence of nodes of each subtree to eventually have a sequence of nodes of a sequence of subtrees. Thus, the transformer-based models are particularly adept at considering the sequential nature of code through the use of positional embeddings and self-attention mechanisms, drawing inspiration from their success in natural language processing tasks. This model utilizes multiple layers of self-attention and feed-forward networks to process data. The model can be mathematically described as follows:

\begin{itemize}
    \item \textbf{Multi-Head Self-Attention:} The self-attention mechanism allows the model to weigh the importance of different tokens within the input sequence differently. This is done using queries ($Q$), keys ($K$), and values ($V$), which are derived from the input embedding matrix $X \in \mathbb{R}^{n \times d}$:

\begin{equation}
    Q = XW^Q, \quad K = XW^K, \quad V = XW^V,
\end{equation}

where $W^Q, W^K, W^V \in \mathbb{R}^{d \times d}$ are parameter matrices. The output of the attention function for a single head is computed as:

\begin{equation}
    \text{Attention}(Q, K, V) = \text{softmax}\left(\frac{QK^T}{\sqrt{d_k}}\right)V,
\end{equation}

where $d_k$ is the dimension of the keys.

In the case of multi-head attention, the above computation is done in parallel for each head, and the outputs are concatenated and linearly transformed:

\begin{equation}
    \text{MultiHead}(Q, K, V) = \text{Concat}(\text{head}_1, \dots, \text{head}_h)W^O,
\end{equation}

where each head is computed as $\text{head}_i = \text{Attention}(QW_i^Q, KW_i^K, VW_i^V)$ and $W^O$ is another parameter matrix.

\item \textbf{Position-wise Feed-Forward Networks:} Each transformer block contains a position-wise feed-forward network, which applies two linear transformations with a ReLU activation in between:

\begin{equation}
    \text{FFN}(x) = \max(0, xW_1 + b_1)W_2 + b_2,
\end{equation}

where $W_1, W_2$ and $b_1, b_2$ are parameters of the layers.

\item \textbf{Layer Normalization and Residual Connections:} Each sub-layer in a transformer, including self-attention and feed-forward networks, has a residual connection around it followed by layer normalization:

\begin{equation}
    \text{LayerNorm}(x + \text{Sublayer}(x)).
\end{equation}

\item \textbf{Output Layer:} The output of the transformer is typically taken from the first token's representation and passed through a final dense layer for classification tasks:

\begin{equation}
    o = \text{softmax}(x_0W + b),
\end{equation}

Where $x_0$ is the transformed embedding of the first token and $W, b$ are parameters of the output layer.

\end{itemize}

\section{Experiment}

\subsection{Experiment Settings}

In this experiment setup, all GNN-based models consist of two convolution layers with hidden dimensions of 40 and 30, followed by two linear layers. To facilitate graph prediction, node representation pooling was employed by concatenating mean and max global pooling techniques. Batch normalization and dropout techniques were applied for training regularization. All models were implemented using PyTorch-Geometric \cite{fey2019fast}.

To standardize our experimental conditions across various utilised models, including TreeCNN, Code2Vec, Transformer-Based, and Dual-Transformer, we trained each for hundred epochs five times with different initialization seeds at a learning rate of \(1 \times 10^{-4}\) and a batch size of four. However, each model requires specific parameters to function optimally. For example, the TreeCNN model uses a node representation embedding size of 100 and a hidden layer size of 300. The Code2Vec model, employing a pre-trained version, was initially trained with 200 distinct contexts and had extensive vocabulary sizes for tokens and paths, set at 1,301,136 and 911,417, respectively, with an embedding size of 128.

We assess both scaled-down (small) and fully scaled versions (large) for the Transformer models. The scaled-down version includes a single transformer block with 768 hidden units, eight self-attention heads, and a maximum sequence length of 2,048 tokens, while the fully scaled version consists of 12 transformer blocks. The implementation of Transformer models utilized the Huggingface library.

By maintaining consistent training epochs, learning rates, and batch size across models and adjusting configurations to meet each model's architectural requirements, our experiment aims to deliver a balanced and comprehensive evaluation of models' performance across various metrics. 










\subsection{Dataset Collection}

In our experiments, to increase reliability, we use two different real-world datasets of performance measurements. The first dataset (\emph{OSSBuild}) is real build data collected from the continuous integration systems of four open-source systems.
The second (\emph{HadoopTests}) is a larger dataset we have collected ourselves by repeatedly executing the unit tests of the Hadoop open-source system in a controlled environment. A summary of both datasets is provided in Table~\ref{tab:dataProjects}.
In the following subsections, we provide some additional information about each of the two datasets that we used in the experimental studies.
\begin{table*}[h!]
\caption{Overview of the OSSBuilds and HadoopTests datasets.}
\label{tab:dataProjects}
\centering
\renewcommand{\arraystretch}{1.3}
\begin{tabular}{l|c|c|c|c|c}
 \textbf{Project} & \textbf{Description} & \textbf{Files} & \textbf{Runs} & \textbf{Nodes} & \textbf{Vocab.} \\
\hline
 sysDS & Apache ML for Data Science lifecycle & 127 & 1321 & 114904 & 3205 \\
 H2 & Java SQL DB & 194 & 1391 & 432375 & 18326 \\
 Dubbo & Apache Remote Procedure Call framework & 123 & 524 & 77142 & 4505 \\
 RDF4J & Scalable RDF & 478 & 1055 & 242673 & 10844 \\
 \textbf{(OSSBuilds) Tot.} & & \textbf{922} & \textbf{4291} & \textbf{867094} & \textbf{36880} \\
\hline
 \textbf{Hadoop} & Apache framework for big data & \textbf{2895} & \textbf{24348} & \textbf{5090798} & \textbf{138952} \\
\hline
\end{tabular}
\end{table*}
\subsubsection{OSSBuild Dataset}
In this dataset (originally used in Samoaa et al.~\cite{SamoaaTEPGNN}), information about test execution times in production build systems was collected for four open-source projects: systemDS, H2, Dubbo, and RDF4J. All four projects use public continuous integration servers containing (public) information about the project's builds, which we harvested for test execution times as a proxy of performance in summer 2021. Basic statistics about the projects in this dataset are described in Table~\ref{tab:dataProjects} (top). "Files" refers to the number of unit test files we collected execution times for, "Runs" is the (total) number of executions of files we extracted data for, whereas "Nodes" and "Vocab." indicate the resulting trees. 
Prior to parsing the test files, we remove code comments to reduce the number of nodes in each tree (by construction, irrelevant). We notice that across 922 ASTs, we have almost 867,000 nodes with 36880 distinct labels as vacabs. 

\subsubsection{HadoopTests Dataset}
To address limitations with the OSSBuilds dataset (primarily the limited number of files for each individual project in the dataset)~\cite{samoaa2023unified}, we additionally collected a second dataset for this study. We selected the Apache Hadoop framework since it entails a large number of test files (2895) of sufficient complexity. We then executed all unit tests in the project five times, recording the execution duration of each test file as reported by the JUnit framework (in millisecond granularity). As an execution environment for this data collection, we used a dedicated virtual machine running in a private cloud environment, with two virtualized CPUs and 8 GByte of RAM. Following performance engineering best practices, we deactivated all other non-essential services while running the tests.
Statistics about the HadoopTests dataset are described in Table~\ref{tab:dataProjects} (bottom).
Since we have more files in HadoopTests, there are more nodes. Thus, ASTs for HadooptTests have almost 5 million nodes and almost 139,000 vocabs.

To better understand our dataset, Table \ref{tab:graphmetrics} shows the average statistics of the input ASTs. In particular, we report the average number of nodes ($|V|$), the average number of edges ($|E|$), the diameter, and density. The data in Table~\ref{tab:graphmetrics} reveals key structural features of the ASTs in our datasets. The near-equal count of nodes ($|V|$) and edges ($|E|$) underscores the tree-like nature of ASTs, where most nodes are directly connected to only one parent. The substantial diameters indicate deep trees, suggesting complex nested code structures. Low-density points to sparse connectivity, emphasizing the depth over breadth in these ASTs, which could affect the performance of neural models that process such hierarchical data.


\begin{table}[h!]
    \caption{Average statistics of the input trees.}
    \label{tab:graphmetrics}
    \centering
    \begin{tabular}{l|ccccc}
         \hline
         Dataset & $|V|$ & $|E|$ & Diameter & Density  \\
         \hline
         {\textbf{OSSBuilds}}& 875 & 874  & 17 & 0.015  \\                               
         \hline
         {\textbf{HadoopTests}}& 1490 & 1489 & 19 & 0.006  \\  
         \hline
    \end{tabular}
\end{table}

\section{ Results}

In this section, we delineate the performance outcomes of our proposed model alongside those of competing frameworks, scrutinized from three distinct angles: initial efficacy across the OssBuilds and Hadoop datasets on a standard data split, variations in model efficiency with increasing sizes of training data, and the adaptability of models through cross-dataset transferability assessments. The evaluation metrics include MSE, MAE, and Pearson correlation coefficient (Cor.), with lower MSE and MAE values indicating better performance and a higher correlation coefficient signifying a stronger relationship between predicted and actual execution times.

\subsection{Models Performance Evaluation}

\begin{table*}[ht]
    \label{tab:generalresults}
    \caption{Test MSE, MAE, and person correlation for both datasets trees}
    \centering
    \begin{tabular}{l|ccc|ccc}
    & \multicolumn{3}{c|}{OssBuilds} & \multicolumn{3}{c}{Hadoop} \\
         \cline{2-7}
    & MSE & MAE & Cor. & MSE & MAE & Cor. \\
    & $\downarrow$ & $\downarrow$ & $\uparrow$ & $\downarrow$ & $\downarrow$ & $\uparrow$ \\
         \hline
    GCN &\mstd{  0.07 }{ 0.02   }&\mstd{  0.21 }{ 0.03   }&\mstd{  0.65 }{ 0.04  }&\mstd{  0.06 }{ 0.02  }&\mstd{  0.21 }{ 0.03   }&\mstd{  0.52 }{ 0.05} \\
     \col{GAT} & \col{\mstd{0.07 }{ 0.01 }}& \col{\mstd{0.23}{ 0.02 }  }& \col{\mstd{0.61  }{ 0.03 }} & \col{\mstd{0.09 }{ 0.01  }}& \col{\mstd{0.25 }{ 0.02 }  }& \col{\mstd{0.38 }{ 0.07 }}\\
     GIN &\mstd{   0.08 }{ 0.01  }&\mstd{  0.21 }{ 0.01  }&\mstd{   0.60 }{ 0.04  }&\mstd{   0.06 }{ 0.003  }&\mstd{  0.20 }{ 0.01   }&\mstd{   0.50 }{ 0.04 }\\
     \col{GraphSage} &\col{\mstd{0.06 }{ 0.01 } }&\col{\mstd{0.20}{0.02}}&\col{\mstd{0.68}{0.02 }}&\col{\mstd{0.06}{0.01}}&\col{\mstd{0.21}{0.02}}&\col{\mstd{0.52 }{ 0.03 }}\\
     \hline
     Code2Vec &\mstd{  0.03 }{ 0.003  }&\mstd{  0.14 }{ 0.01  }&\mstd{  0.44 }{ 0.11   }&\mstd{  0.03 }{ 0.002  }&\mstd{  0.14 }{ 0.01  }&\mstd{  0.28 }{ 0.05 }\\
     \col{TreeCNN} &\col{\mstd{  0.03 }{ 0.002 } }&\col{\mstd{  0.13 }{ 0.004 }}&\col{\mstd{   0.51 }{ 0.03 }  }&\col{\mstdb{  0.02 }{ 0.001 } }&\col{\mstd{  0.12 }{ 0.01 } }&\col{\mstd{  0.57 }{ 0.03 }}\\
     Transformer-Based (small) &\mstd{  0.09 }{ 0.04  }&\mstd{  0.25 }{ 0.07  }&\mstd{  0.45 }{ 0.15   }&\mstd{  0.08 }{ 0.0.03  }&\mstd{  0.24 }{ 0.06  }&\mstd{  0.27 }{ 0.06 }\\
     \col{Transformer-Based (large)} &\col{\mstd{ 0.46 }{ 0.25 } }&\col{\mstd{ 0.56 }{ 0.20 } }&\col{\mstd{ 0.30 }{ 0.10 }  }&\col{\mstd{ 0.05 }{ 0.01 } }&\col{\mstd{ 0.17 }{ 0.03 } }&\col{\mstd{ 0.40 }{ 0.07 }}\\
     
     \hline
     \textbf{Dual-Transformer (small)} &\mstdb{  0.01 }{ 0.002  }&\mstdb{  0.08 }{ 0.006  }&\mstdb{  0.85 }{ 0.02 }&\mstdb{  0.02 }{ 0.01  }&\mstd{  0.12 }{ 0.03  }&\mstd{  0.67 }{ 0.04 }\\
     \col{\textbf{Dual-Transformer (large)}} & \col{\mstd{0.02}{0.006}} & \col{\mstd{0.09}{0.02}} & \col{\mstd{0.83}{0.03}}&  \col{\mstdb{0.02}{0.004}} & \col{\mstdb{0.11}{0.01}} & \col{\mstdb{0.72}{0.03}} \\
     \hline
    \end{tabular} 
\end{table*}
\begin{figure*}
    \centering
    \includegraphics[width=\linewidth]{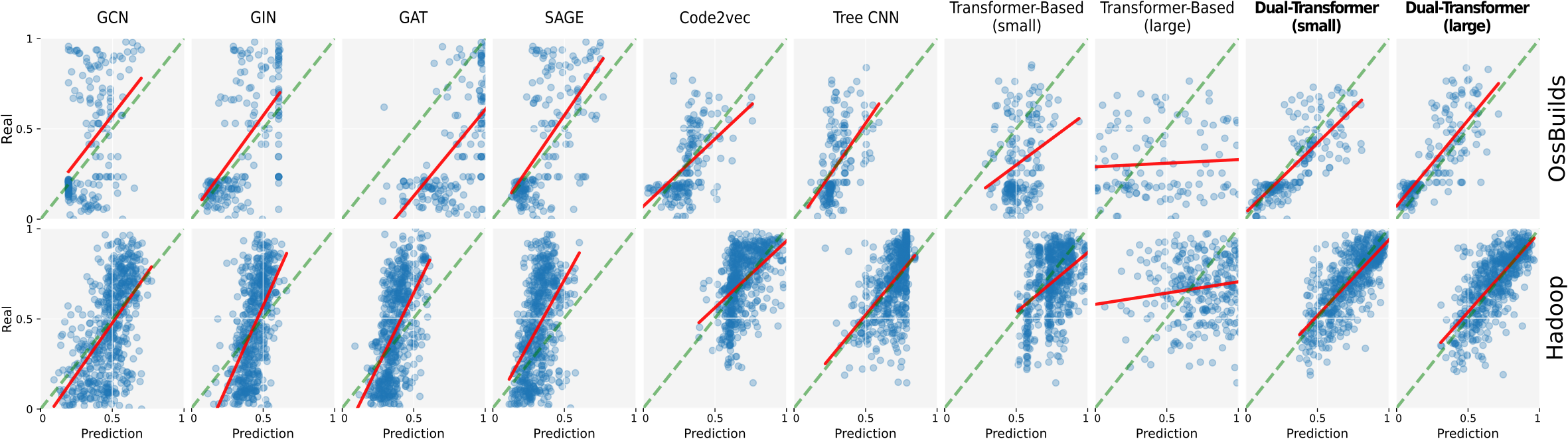}
    \caption{\textbf{Real vs predicted values} Each panel reports the real (y-axes) and predicted (x-axes) values for each model. Each pair that is real-predicted is represented as a blue point, while the dashed red line shows a linear regression model fitted to the data.}
    \label{fig:correlation}
\end{figure*}

This section presents a comprehensive analysis of our experimental results, comparing the performance of various models on the standard split of datasets. Thus, for five different seeds, we split the data into 80\% used for training and the rest 20\% for testing, and then we report the average results across the seeds.
\paragraph{GNN models}
As presented in Table~\ref{tab:generalresults}, among the GNN architectures, GraphSage exhibits superior performance on the OssBuilds dataset, with the lowest MSE of 0.06 and MAE of 0.20, alongside a commendable correlation of 0.68. However, on the Hadoop dataset, all GNN models demonstrate relatively similar MSE scores of 0.06 (except for GAT). As for MAE and Correlation scores, both GCN and GraphSage have similar scores, 0.21 and 0.52, respectively, with GraphSage maintaining a slight edge in standard deviation(STD). It is worth mentioning that the GIN model performs well for the Hadoop dataset with the best MAE score. 

\paragraph{{TBNN Models}}
Transitioning to the TBNN competitors, TreeCNN TreeCNN emerges as a strong contender in all metrics (particularly in Pearson correlation), showcasing the best scores among TBNN models for both datasets, with the lowest MSE of 0.03, an MAE of 0.13 and the highest correlation score of 0.57 on the OssBuilds dataset and 0.02, 0.12, and 0.57 for MSE, MAE, and Pearson correlation On the Hadoop dataset. It is worth mentioning that Code2Vec is a good competitor with a very small margin of difference from TreeCNN in both datasets, especially for error metrics.  As for transformer-based, it achieves the worst results, especially in the large setting on the OssBuilding dataset, explaining that the model is too complex for this dataset. In contrast, the efficiency of the same model with the same setting is improving in the Hadoop dataset, which is the larger dataset with more complex trees.

\paragraph{{Our Dual-Transformer Model}}
Our proposed Dual-Transformer model (in both large and small settings) significantly outperforms both the GNN architectures and TBNN competitors across all metrics on both datasets. It achieves the best MSE of 0.01 and 0.02 and an outstanding MAE of 0.08 and 0.09 on the OssBuilds dataset, coupled with a remarkable correlation of 0.85 and 0.83 for small and large settings, respectively. Although the Hadoop dataset shares the best MSE score of 0.02 with TreeCNN, its MAE of 0.12 and 0.11, as well as the correlation of 0.67 and 0.72, is unparalleled, underscoring its superior predictive capability and efficiency in capturing the underlying complexities of source code.

\paragraph{{Models' Prediction Trending Analysis}}
In Figure \ref{fig:correlation}, a correlation chart is presented for the OssBuilds dataset, each panel depicting a distinct model. The figure clearly indicates that the GNN-based models struggle to predict real values accurately. Notably, it appears that all GNN-based models exhibit a tendency to predict values that are either close to one or close to zero. The Transformer-based, TreeCNN, and Code2Vec models encounter challenges in predicting larger values. While our model also faces difficulty in predicting larger values, however, it outperforms the other approaches.

\paragraph{{Conclusion}}

These results in Table~\ref{tab:generalresults} underscore the efficacy of our Dual-Transformer approach, particularly in its ability to harness the syntactic and semantic intricacies of ASTs for source code analysis. The significant improvement in correlation coefficients highlights the model's adeptness at understanding the nuanced relationships within the code, making it a promising tool for developers seeking early insights into the potential execution characteristics of their programs.
When compared with transformer-based models, our dual transformer has displayed superior performance, which can be attributed to its specialized architecture designed to handle dual input modalities. On the other hand, the Transformer-based model may not effectively capture the interplay between different types of input data, such as textual and structural representations in programming code. Furthermore, the transformer-based model may have limitations in dealing with extended sequence lengths compared to the dual transformer.

GNN models seem to be the second-best regarding Pearson scores for both datasets ( except for TreeCNN in Hadoop). However, both error metrics for code2vec and TreeCNN models are better for both datasets compared with GNN models. 

Regarding trending results for the models between OssBuilding and Hadoop, only the TreeCNN model has an upward trend. In contrast, the scores for the rest of the models have decreased in the Hadoop dataset compared to OssBuilding. 

In conclusion, the empirical evidence strongly supports adopting our dual transformer model for the regression task. By effectively leveraging both the lexical and syntactic features of the source code, our approach establishes a new state-of-the-art performance, paving the way for future research in this domain.

\subsection{Assessing Model Efficiency Across Incremental Training Data Sizes}

Given the expense associated with data collection, this section delves into the effectiveness of our and other models when trained on a reduced dataset. 
Specifically, we employ random selection to allocate 20$\%$, 40$\%$, and 60$\%$ of the dataset for training the models, while a fixed 20\% of the remaining portion is designated for testing. 
Through the outcomes shown in Table~\ref{tab:extended_training_data_sizes}, we observed distinct patterns of performance adaptation as the models were exposed to increasing proportions of the dataset.
\paragraph{{GNN models}}
For the OssBuilds dataset, GNN models generally demonstrated an expected trend: improvements in MSE(except for GIN), MAE (except for GIN and GraphSage), and Pearson correlation (except GCN, GIN) as the training data size expanded from 20\% to 60\%.  Except for GIN, all other GNN models have stability in MSE score in the training data size of 40\% to 60\%. Moreover, the GAT model also has stability in MAE for both 20\% and 40\% of training data.

For the Hadoop dataset, GNN models tend to have better error metrics (except for GAT) across all sizes used in training. However, the opposite is true when it comes to Pearson's score. Similar to OssBuids, there is some stability in error metrics scores as in GIN in 20\% and 40\% and GraphSage in 40\% and 60\% sizes of training data. As for GCN, the MAE is stable across all training seizes. As for GAT, we got the same MSE score when we used 20\% and 40\% for training.
Despite the stability in error metrics most of the time, the prediction correlation of Pearson correlation still changes through the different sizes, which justifies the importance of adding more data for training. Thus, notably, GNN-based models benefit from more extensive data to better capture the structural nuances of the ASTs. However, their performance plateaus, suggesting a limit to how much simply increasing data can benefit these models without corresponding adjustments in model complexity or architecture

\paragraph{{TBNN Models}}

For the OssBuilds dataset, the TBNN models show varying degrees of sensitivity to the amount of training data. Code2Vec and TreeCNN exhibit stable performance in terms of MSE and MAE across different training sizes, but Pearson correlation shows a slight increase, indicating better alignment between predicted and actual values with more data. Transformer-based models, both small and large, display inconsistent performance, with significant fluctuations in MSE and MAE and only modest improvements in Pearson correlation, suggesting instability in their learning process with varied data sizes.

In the Hadoop dataset, TBNN models follow a similar trend. Code2Vec and TreeCNN maintain stable error metrics, while the Pearson correlation improves as more training data is provided. The Transformer-Based models show less stability, with large variations in MSE, MAE, and Pearson correlation across different training sizes. This instability indicates that Transformer-Based models might require more fine-tuning or adjustments in their architecture to better handle varied data sizes.

\paragraph{{Our Dual-Transformer Model}}

The Dual-Transformer models, our proposed approach, show robust performance improvements across all metrics as the training data size increases. For the OssBuilds dataset, both small and large variants of the Dual-Transformer model exhibit significant decreases in MSE and MAE and substantial increases in Pearson correlation, reaching 0.82 with 60\% training data. This indicates the model's ability to effectively utilize additional training data to enhance predictive accuracy and correlation alignment.

Similarly, in the Hadoop dataset, the Dual-Transformer models demonstrate consistent performance gains. The MSE and MAE decrease steadily, and the Pearson correlation shows marked improvement, reaching 0.73 with 60\% training data. This consistent improvement across different data sizes underscores the effectiveness of the Dual-Transformer architecture in handling complex regression tasks on ASTs, leveraging the added data to refine its predictive capabilities.
\paragraph{{Conclusion}}

The analysis across incremental training data sizes reveals that while GNN models benefit from increased data, their performance gains plateau. TBNN models show varying degrees of stability, with Code2Vec and TreeCNN being more resilient to changes in data size.

The overall error values for the Hadoop dataset were generally smaller than the OssBuilds dataset (except for GAT). On the other hand, Pearson Correlation scores are better for OssBuilds for all models except TreeCNN, suggesting that Hadoop's complexity and tree structures might pose additional challenges in prediction. Despite this, the Dual-Transformer model consistently outperformed other models, reaffirming its adaptability and efficiency in handling complex trees. In addition, it demonstrates robust scalability and effectiveness in leveraging additional data to enhance predictive accuracy and correlation, making them well-suited for regression tasks for trees. It is worth mentioning that the Dual-Transformer model can be satisfied with 40\% of training data to achieve the best performance when it comes to the Hadoop dataset. Still, it consistently needs more data to be utilised in training in OssBuilds, which contains samples from 4 different projects. 

These findings underscore the critical importance of data volume in training neural network models for source code analysis. They also highlight the Dual-Transformer model's superiority in adapting to varied training sizes while maintaining robust performance, marking it as a promising approach for efficient source code analysis.

\begin{table*}[ht]
    \centering
    \caption{Test MSE, MAE, and Pearson correlation across different training data sizes for both datasets}
    \label{tab:extended_training_data_sizes}
    \resizebox{\textwidth}{!}{%
    \setlength{\tabcolsep}{3pt} 
    \begin{tabular}{l|ccc|ccc|ccc}
    \hline
    & \multicolumn{9}{c}{OssBuilds} \\
    \cline{2-10}
    & \multicolumn{3}{c|}{20\%} & \multicolumn{3}{c|}{40\%} & \multicolumn{3}{c}{60\%} \\
    Model & MSE & MAE & Cor. & MSE & MAE & Cor. & MSE & MAE & Cor. \\
    \hline
    GCN & \mstd{0.09}{0.02} & \mstd{0.26}{0.03} & \mstd{0.51}{0.07} & \mstd{0.08}{0.01} & \mstd{0.23}{0.03} & \mstd{0.63}{0.03} & \mstd{0.08}{0.01} & \mstd{0.22}{0.02} & \mstd{0.62}{0.05}  \\
    \col{GAT} & \col{\mstd{0.08}{0.005}} & \col{\mstd{0.23}{0.02}} & \col{\mstd{0.52}{0.04}} & \col{\mstd{0.07}{0.005}} & \col{\mstd{0.23}{0.02}} & \col{\mstd{0.57}{0.02}} & \col{\mstd{0.07}{0.004}} & \col{\mstd{0.22}{0.02}} & \col{\mstd{0.59}{0.04}} \\
    GIN & \mstd{0.09}{0.01} & \mstd{0.24}{0.02} & \mstd{0.49}{0.06} & \mstd{0.07}{0.01} & \mstd{0.21}{0.009} & \mstd{0.57}{0.05} & \mstd{0.09}{0.02} & \mstd{0.24}{0.04} & \mstd{0.56}{0.08} \\
    \col{GraphSage} & \col{\mstd{0.08}{0.007}} & \col{\mstd{0.24}{0.02}} & \col{\mstd{0.58}{0.005}} & \col{\mstd{0.07}{0.01}} & \col{\mstd{0.22}{0.03}} & \col{\mstd{0.64}{0.03}} & \col{\mstd{0.07}{0.01}} & \col{\mstd{0.23}{0.03}} & \col{\mstd{0.65}{0.04}} \\
    \hline
    Code2Vec & \mstd{0.03}{0.002} & \mstd{0.15}{0.006} & \mstd{0.33}{0.08} & \mstd{0.03}{0.002} & \mstd{0.15}{0.01} & \mstd{0.35}{0.11} & \mstd{0.03}{0.002} & \mstd{0.15}{0.01} & \mstd{0.39}{0.07} \\
    \col{TreeCNN} & \col{\mstd{0.03}{0.001}} & \col{\mstd{0.15}{0.005}} & \col{\mstd{0.32}{0.06}} & \col{\mstd{0.03}{0.001}} & \col{\mstd{0.15}{0.003}} & \col{\mstd{0.40}{0.05}} & \col{\mstd{0.03}{0.003}} & \col{\mstd{0.14}{0.01}} & \col{\mstd{0.46}{0.05}} \\
    Transformer-Based (small) & \col{\mstd{0.13}{0.04}} & \col{\mstd{0.29}{0.05}} & \col{\mstd{0.37}{0.08}} & \col{\mstd{0.14}{0.09}} & \col{\mstd{0.30}{0.10}} & \col{\mstd{0.39}{0.12}} & \col{\mstd{0.09}{0.05}} & \col{\mstd{0.25}{0.07}} & \col{\mstd{0.34}{0.22}} \\
    \col{Transformer-Based (large)} & \mstd{0.56}{0.47} & \mstd{0.61}{0.17} & \mstd{0.18}{0.11} & \mstd{0.60}{0.26} & \mstd{0.62}{0.14} & \mstd{0.13}{0.12} & \mstd{0.34}{0.06} & \mstd{0.45}{0.03} & \mstd{0.26}{0.05}  \\
    \hline
    \textbf{Dual-Transformer (small)} & \col{\mstdb{0.02}{0.004}} & \col{\mstdb{0.10}{0.01}} & \col{\mstdb{0.79}{0.04}} & \col{\mstdb{0.01}{0.003}} & \col{\mstdb{0.09}{0.007}} & \col{\mstdb{0.84}{0.03}} & \col{\mstd{0.02}{0.01}} & \col{\mstd{0.10}{0.03}} & \col{\mstdb{0.82}{0.05}} \\
    \col{\textbf{Dual-Transformer (large)}} & \mstdb{0.02}{0.004} & \mstd{0.11}{0.01} & \mstd{0.71}{0.04} & \mstd{0.02}{0.002} & \mstdb{0.09}{0.005} & \mstd{0.79}{0.05} & \mstdb{0.01}{0.002} & \mstdb{0.09}{0.01} & \mstdb{0.82}{0.02}
    \\ \hline
    & \multicolumn{9}{c}{Hadoop} \\
    \cline{2-10}
    & \multicolumn{3}{c|}{20\%} & \multicolumn{3}{c|}{40\%} & \multicolumn{3}{c}{60\%} \\
    Model & MSE & MAE & Cor. & MSE & MAE & Cor. & MSE & MAE & Cor. \\
    \hline
    GCN & \mstd{0.07}{0.01} & \mstd{0.22}{0.02} & \mstd{0.42}{0.12} & \mstd{0.08}{0.02} & \mstd{0.22}{0.03} & \mstd{0.50}{0.03} & \mstd{0.07}{0.01} & \mstd{0.22}{0.02} & \mstd{0.52}{0.06} \\
    \col{GAT} & \col{\mstd{0.09}{0.02}} & \col{\mstd{0.25}{0.02}} & \col{\mstd{0.17}{0.17}} & \col{\mstd{0.09}{0.009}} & \col{\mstd{0.24}{0.1}} & \col{\mstd{0.19}{0.09}} & \col{\mstd{0.11}{0.04}} & \col{\mstd{0.27}{0.04}} & \col{\mstd{0.17}{0.22}} \\
    GIN & \mstd{0.07}{0.02} & \mstd{0.22}{0.03} & \mstd{0.36}{0.14} & \mstd{0.07}{0.01} & \mstd{0.22}{0.02} & \mstd{0.46}{0.06} & \mstd{0.06}{0.006} & \mstd{0.20}{0.009} & \mstd{0.48}{0.09} \\
    \col{GraphSage} & \col{\mstd{0.07}{0.003}} & \col{\mstd{0.22}{0.009}} & \col{\mstd{0.35}{0.08}} & \col{\mstd{0.06}{0.008}} & \col{\mstd{0.21}{0.02}} & \col{\mstd{0.57}{0.09}} & \col{\mstd{0.06}{0.01}} & \col{\mstd{0.21}{0.02}} & \col{\mstd{0.55}{0.08}} \\
    \hline
    Code2Vec & \mstd{0.04}{0.003} & \mstd{0.13}{0.006} & \mstd{0.38}{0.04} & \mstd{0.03}{0.003} & \mstd{0.14}{0.01} & \mstd{0.24}{0.04} & \mstdb{0.03}{0.001} & \mstd{0.13}{0.004} & \mstd{0.38}{0.06} \\
    \col{TreeCNN} & \col{\mstdb{0.02}{0.003}} & \col{\mstd{0.12}{0.002}} & \col{\mstd{0.44}{0.03}} & \col{\mstd{0.02}{0.001}} & \col{\mstd{0.12}{0.006}} & \col{\mstd{0.51}{0.02}} & \col{\mstdb{0.02}{0.001}} & \col{\mstd{0.11}{0.001}} & \col{\mstd{0.53}{0.01}} \\
    Transformer-Based (small) & \col{\mstd{0.09}{0.05}} & \col{\mstd{0.23}{0.08}} & \col{\mstd{0.26}{0.16}} & \col{\mstd{0.10}{0.04}} & \col{\mstd{0.26}{0.07}} & \col{\mstd{0.34}{0.11}} & \col{\mstd{0.07}{0.04}} & \col{\mstd{0.21}{0.07}} & \col{\mstd{0.38}{0.17}}  \\
    \col{Transformer-Based (large)} &  \mstd{0.28}{0.09} & \mstd{0.42}{0.07} & \mstd{0.14}{0.05} & \mstd{0.08}{0.01} & \mstd{0.22}{0.02} & \mstd{0.29}{0.07} & \mstd{0.09}{0.06} & \mstd{0.24}{0.10} & \mstd{0.34}{0.02}  \\
    \hline
    Dual-Transformer (small) & \col{\mstdb{0.02}{0.002}} & \col{\mstdb{0.11}{0.008}} & \col{\mstdb{0.70}{0.04}} & \col{\mstdb{0.02}{0.005}} & \col{\mstd{0.11}{0.02}} & \col{\mstdb{0.72}{0.03}} & \col{\mstdb{0.02}{0.005}} & \col{\mstd{0.11}{0.02}} & \col{\mstd{0.68}{0.02}}   \\
    \col{Dual-Transformer (large)} & \mstdb{0.02}{0.007} & \mstd{0.13}{0.02} & \mstd{0.67}{0.03} & \mstdb{0.02}{0.002} & \mstdb{0.10}{0.007} & \mstdb{0.72}{0.01} & \mstdb{0.02}{0.002} & \mstdb{0.09}{0.008} & \mstdb{0.73}{0.01} \\
    \hline
    \end{tabular}}
\end{table*}

\subsection{Cross-Dataset Transferability}\label{subsec:tran}
In this section, we explore the models' performance in an inductive scenario, where they are trained on one dataset and tested on another dataset~\cite{longa2023graph}. The results, shown in Table~\ref{tab:transferability}, unfold distinct patterns of performance across the GNN-based models, TBNNs, and our Dual-Transformer model.
Due to the differing scales of the machines used to collect the datasets (since the hardware used is one of the factors that affect the execution time), it was imperative to adapt the model to each specific context. To this end, we initially trained the model on one dataset and subsequently fine-tuned it using a small subset of the other dataset to optimize its parameters. This fine-tuning process involved using incremental portions of the test dataset—specifically 10\%, 20\%, and 30\%—to refine the model's ability to generalize across different operational conditions. The efficacy of the fine-tuning was then evaluated by testing the model on 20\% of the test dataset, which is fixed across all portions, ensuring a consistent assessment framework across all experimental conditions.
This methodological approach allowed us to rigorously assess the model's adaptability and performance across datasets characterized by diverse computational environments.

When we \textbf{trained the models on Hadoop and fine-tuned and evaluated on Ossbuilds}, models exhibited relatively stable MSE and MAE across all fine-tuning portions (except for the small version of our dual transformer model), indicating a capacity to maintain consistent error rates when transferring knowledge from a larger (Hadoop) to a smaller dataset (OssBuilds). Pearson correlation showed gradual improvement as the fine-tuning portion increased (except for a small model of transformed-based), highlighting a modest but positive adaptation. 
The Dual-Transformer models demonstrated the best performance, with significant improvements in correlation and error rates, making them the most adaptable across dataset sizes. Since the error metrics are slightly better for the large version of our dual transformer, the prediction correlation score for the small version, however, is largely better across the usage of all potions of fine-tuning. 
Conversely, the Transformer-Based models struggled the most, especially in larger configurations, showing limited correlation improvements and variable error rates, suggesting challenges in adapting to the smaller dataset's nuances.
It is worth mentioning that GNN models show more efficient predictions since the Pearson correlation score is better than all other TBNN models. However, this is also the case regarding the error metrics.
Compared to other experiments,  TBNN, we see a huge decrease in error metrics compared to the previous experiments. However, the error metrics of GNN are somehow stable across all our experiments. As for our model, only the MAE metric increased.

In the scenario where \textbf{OssBuilds was used for training and Hadoop for fine-tuning and evaluation}, all models generally showed better MSE and MAE scores compared to the previous setting, especially for TBNN models since both code2vec and TreeCNN competing our model, particularly with a small portion of test data used for fine-tuning. That said, when we train TBNN models on trees for different projects, the models can easily generalized to other trees. However, Pearson correlation is still an issue for these models. 
GNN models have slightly similar error metrics; however, the correlation score is decreasing ( except for GAT). That is reasonable since GNN models mainly learn based on the tree's structure throughout collecting information from the neighbour nodes. 
Although transformer-based is still the worst model in terms of Pearson score, the error metrics have improved hugely, especially for the small version of the model, which puts this version in a better position compared to GNN. The large version of transformer-based is generally still the worst in all metrics. 
Our model is still the best model, even in this scenario, with significantly better errors and correlation scores than Hadoop's usage in the training. 
We also observe in this scenario that with increased fine-tuning, GraphSage and TreeCNN are stable across the portions. 
The improvement in the models in this scenario shows that when we train the models on diverse trees( since OssBuild contains four projects with four different trees), we can have a better generalization compared to training the models on a large but not diverse tree ( as in Hadoop, where all trees come from one project).

\begin{table*}[ht]
    \centering
    \caption{Transferability Test MSE, MAE, and Pearson correlation for both datasets}
    \label{tab:transferability}
    \resizebox{\textwidth}{!}{%
    \setlength{\tabcolsep}{3pt} 
    \begin{tabular}{l|ccc|ccc|ccc}
    \hline
    & \multicolumn{9}{c}{Train Set = Hadoop \& Test Set = OssBuild} \\
    \cline{2-10}
    & \multicolumn{3}{c|}{10\%} & \multicolumn{3}{c|}{20\%} & \multicolumn{3}{c}{30\%} \\
    Model & MSE & MAE & Cor. & MSE & MAE & Cor. & MSE & MAE & Cor. \\
    \hline
    GCN & \mstd{0.08}{0.01} & \mstd{0.23}{0.02} & \mstd{0.47}{0.02} & \mstd{0.08}{0.01} & \mstd{0.22}{0.03} & \mstd{0.51}{0.02} & \mstd{0.08}{0.01} & \mstd{0.22}{0.03} & \mstd{0.53}{0.02}  \\
    \col{GAT} & \col{\mstd{0.09}{0.01}} & \col{\mstd{0.26}{0.02}} & \col{\mstd{0.35}{0.20}} & \col{\mstd{0.08}{0.01}} & \col{\mstd{0.24}{0.03}} & \col{\mstd{0.40}{0.21}} & \col{\mstd{0.08}{0.01}} & \col{\mstd{0.24}{0.03}} & \col{\mstd{0.41}{0.19}} \\
    GIN & \mstd{0.08}{0.01} & \mstd{0.24}{0.02} & \mstd{0.51}{0.02} & \mstd{0.08}{0.01} & \mstd{0.22}{0.02} & \mstd{0.54}{0.02} & \mstd{0.07}{0.01} & \mstd{0.21}{0.02} & \mstd{0.55}{0.02} \\
    \col{GraphSage} & \col{\mstd{0.08}{0.01}} & \col{\mstd{0.25}{0.02}} & \col{\mstd{0.51}{0.02}} & \col{\mstd{0.08}{0.01}} & \col{\mstd{0.23}{0.03}} & \col{\mstd{0.55}{0.02}} & \col{\mstd{0.08}{0.01}} & \col{\mstd{0.23}{0.03}} & \col{\mstd{0.56}{0.02}} \\
    \hline
    Code2Vec & \mstd{0.20}{0.01} & \mstd{0.37}{0.02} & \mstd{0.21}{0.09} & \mstd{0.20}{0.01} & \mstd{0.37}{0.02} & \mstd{0.30}{0.10} & \mstd{0.19}{0.02} & \mstd{0.36}{0.02} & \mstd{0.33}{0.09} \\
    \col{TreeCNN} & \col{\mstd{0.21}{0.01}} & \col{\mstd{0.39}{0.01}} & \col{\mstd{0.14}{0.07}} & \col{\mstd{0.20}{0.005}} & \col{\mstd{0.38}{0.01}} & \col{\mstd{0.25}{0.05}} & \col{\mstd{0.19}{0.005}} & \col{\mstd{0.37}{0.01}} & \col{\mstd{0.32}{0.03}} \\
    Transformer-Based (small) & \mstd{0.24}{0.07} & \mstd{0.39}{0.05} & \mstd{0.40}{0.10} & \mstd{0.34}{0.13} & \mstd{0.44}{0.10} & \mstd{0.40}{0.17} & \mstd{0.24}{0.03} & \mstd{0.40}{0.04} & \mstd{0.35}{0.15} \\
    \col{Transformer-Based (large)} & \col{\mstd{0.64}{0.35}} & \col{\mstd{0.90}{0.60}} & \col{\mstd{0.02}{0.09}} & \col{\mstd{0.88}{0.48}} & \col{\mstd{0.82}{0.04}} & \col{\mstd{0.07}{0.04}} & \col{\mstd{0.42}{0.78}} & \col{\mstd{0.68}{0.41}} & \col{\mstd{0.15}{0.07}} \\
    \hline
    \textbf{Dual-Transformer (small)} & \mstd{0.11}{0.02} & \mstd{0.26}{0.02} & \mstdb{0.47}{0.07} & \mstd{0.08}{0.02} & \mstd{0.23}{0.02} & \mstdb{0.56}{0.10} & \mstd{0.06}{0.01} & \mstd{0.20}{0.01} & \mstdb{0.65}{0.09} \\
    \col{\textbf{Dual-Transformer (large)}} & \col{\mstdb{0.05}{0.005}} & \col{\mstdb{0.17}{0.006}} & \col{\mstd{0.15}{0.05}} & \col{\mstdb{0.04}{0.006}} & \col{\mstdb{0.16}{0.01}} & \col{\mstd{0.23}{0.04}} & \col{\mstdb{0.03}{0.01}} & \col{\mstdb{0.15}{0.04}} & \col{\mstd{0.42}{0.06}} \\ \hline
    & \multicolumn{9}{c}{Train Set = OssBuilds \& Test Set = Hadoop} \\
    \cline{2-10}
    & \multicolumn{3}{c|}{10\%} & \multicolumn{3}{c|}{20\%} & \multicolumn{3}{c}{30\%} \\
    Model & MSE & MAE & Cor. & MSE & MAE & Cor. & MSE & MAE & Cor. \\
    \hline
    GCN & \mstd{0.07}{0.01} & \mstd{0.22}{0.01} & \mstd{0.43}{0.05} & \mstd{0.07}{0.01} & \mstd{0.21}{0.01} & \mstd{0.46}{0.04} & \mstd{0.09}{0.02} & \mstd{0.24}{0.03} & \mstd{0.44}{0.05} \\
    \col{GAT} & \col{\mstd{0.10}{0.01}} & \col{\mstd{0.25}{0.01}} & \col{\mstd{0.37}{0.03}} & \col{\mstd{0.09}{0.01}} & \col{\mstd{0.24}{0.1}} & \col{\mstd{0.43}{0.02}} & \col{\mstd{0.09}{0.01}} & \col{\mstd{0.25}{0.02}} & \col{\mstd{0.44}{0.01}} \\
    GIN & \mstd{0.09}{0.03} & \mstd{0.24}{0.03} & \mstd{0.42}{0.06} & \mstd{0.08}{0.02} & \mstd{0.23}{0.03} & \mstd{0.46}{0.05} & \mstd{0.07}{0.02} & \mstd{0.22}{0.02} & \mstd{0.47}{0.04} \\
    \col{GraphSage} & \col{\mstd{0.07}{0.01}} & \col{\mstd{0.22}{0.01}} & \col{\mstd{0.45}{0.04}} & \col{\mstd{0.07}{0.01}} & \col{\mstd{0.22}{0.01}} & \col{\mstd{0.48}{0.03}} & \col{\mstd{0.06}{0.004}} & \col{\mstd{0.21}{0.005}} & \col{\mstd{0.52}{0.02}} \\
    \hline
    Code2Vec & \mstdb{0.005}{0.0005} & \mstdb{0.06}{0.003} & \mstd{0.16}{0.07} & \mstd{0.004}{0.002} & \mstd{0.06}{0.005} & \mstd{0.38}{0.03} & \mstd{0.005}{0.0002} & \mstd{0.06}{0.004} & \mstd{0.38}{0.05} \\
    \col{TreeCNN} & \col{\mstdb{0.005}{0.0004}} & \col{\mstdb{0.06}{0.003}} & \col{\mstd{0.30}{0.04}} & \col{\mstd{0.005}{0.001}} & \col{\mstd{0.06}{0.004}} & \col{\mstd{0.38}{0.03}} & \col{\mstd{0.004}{0.0003}} & \col{\mstdb{0.05}{0.001}} & \col{\mstd{0.43}{0.02}} \\
    Transformer-Based (small) & \mstd{0.04}{0.01} & \mstd{0.16}{0.03} & \mstd{0.04}{0.09} & \mstd{0.08}{0.05} & \mstd{0.23}{0.09} & \mstd{0.11}{0.12} & \mstd{0.04}{0.03} & \mstd{0.16}{0.07} & \mstd{0.14}{0.07} \\
    \col{Transformer-Based (large)} & \col{\mstd{0.26}{0.02}} & \col{\mstd{0.41}{0.02}} & \col{\mstd{0.11}{0.06}} & \col{\mstd{0.15}{0.08}} & \col{\mstd{0.31}{0.10}} & \col{\mstd{0.06}{0.02}} & \col{\mstd{0.14}{0.04}} & \col{\mstd{0.31}{0.05}} & \col{\mstd{0.12}{0.07}} \\
    \hline
    Dual-Transformer (small) & \mstd{0.006}{0.005} & \mstdb{0.06}{0.03} & \mstdb{0.62}{0.03} & \mstd{0.004}{0.001} & \mstd{0.05}{0.006} & \mstd{0.64}{0.05} & \mstd{0.004}{0.001} & \mstdb{0.05}{0.006} & \mstd{0.61}{0.06} \\
    \col{Dual-Transformer (large)} & \col{\mstdb{0.005}{0.002}} & \col{\mstdb{0.06}{0.01}} & \col{\mstd{0.56}{0.09}} & \col{\mstdb{0.003}{0.002}} & \col{\mstdb{0.04}{0.001}} & \col{\mstdb{0.67}{0.03}} & \col{\mstdb{0.003}{0.0004}} & \col{\mstdb{0.05}{0.004}} & \col{\mstdb{0.67}{0.03}} \\
    \hline
    \end{tabular}}
\end{table*}

\section{Discussion}


The analysis of our experimental results, particularly focusing on the Pearson correlation coefficients and error metrics (MSE and MAE), reveals insightful trends in the performance of various models applied to trees. Remarkably, our model demonstrates superior performance compared to other methods in all preceding experiments. 

Based on Sections~\ref{sec:dualtransformer}, \ref{sec:othermodels}, we can differentiate the learning of the models into two categories: 1) GNNs and TreeCNN, which learn based on the structure of the tree without any interest in the tokens of the nodes. 2) The rest of the models use the sequence of tokens of the nodes for learning. 
The first category of models looks at the topological structure of the tree. Whereas the second category focuses on the exact sequence of tokens of the tree node (e.g., class definition, control statement, method declaration, etc.)
Thus, in this section, we will comment on the results we observed in the previous section.  
\subsection{Our transformer model vs the competitor}
The Dual-Transformer's design, which incorporates cross-attention between the token-level transformer and the tree node-level transformer, allows for a richer representation of the source code by highlighting the interaction between lexical and syntactic features. This nuanced representation is likely the reason for the observed improvement in performance. In contrast, the baseline Transformer-Based model, which operates solely on the tree, does not capture the lexical context to the same extent, thereby limiting its effectiveness for the regression task.

\subsection{GNN part}
GNN-based approaches consistently underperform compared to our model. The failure of GNNs in learning meaningful representations is attributed to the inherent topological structure of trees. Table~\ref{tab:graphmetrics} reveals that the average network diameter is 17 and 19 for OssBuilds and Hadoop, respectively. The elevated diameter poses a significant challenge for GNN-based methods, necessitating deeper networks. However, this exacerbates well-known issues such as over-smoothing\cite{rusch2023survey} and over-squashing\cite{alon2021on}. 

\subsection{Attention mechanism}
The application of attention mechanisms across different models unveils varied outcomes. Our Dual-Transformer model, leveraging cross-attention, consistently outperforms other approaches, underscoring the efficacy of attention in distilling relevant features from both token sequences and AST node sequences. In contrast, the GAT model exhibits a mixed performance, ranking well on the OssBuilds dataset but falling short on the Hadoop dataset. This inconsistency highlights the potential sensitivity of attention-based GNNs to the underlying dataset characteristics. Interestingly, the TreeCNN model, which does not employ attention, demonstrates resilience across datasets, suggesting that attention mechanisms, while powerful, are not a panacea and may introduce complexity that does not always translate to improved performance.

\subsection{Level of Inforamtion}
The distinction between node-level and path-level information processing is another critical factor in our analysis. Except for code2vec, which operates at the path level, all other models process information at the node level. This distinction might contribute to the unique positioning of code2vec in the performance spectrum, indicating that the granularity of analysis (node vs. path) can significantly influence model outcomes.

\subsection{Error Analysis and Pearson correlation score}
It is noteworthy to highlight that the MAE, MSE, and Pearson correlation capture distinct aspects. To elaborate, MAE offers an assessment of error magnitude, disregarding their direction. Conversely, MSE accentuates larger errors through the squaring operation, rendering it sensitive to outliers. Lastly, Pearson correlation gauges the linear relationship between two variables, providing a measure of the strength and direction of the linear association between predicted and actual values.
\begin{enumerate}

 \item \textbf{Mean Squared Error (MSE) and Mean Absolute Error (MAE)} primarily measure the accuracy of predictions in terms of error magnitude. Both metrics are direct measures of the average errors made by the model:
\begin{itemize}
    \item \textbf{MSE} gives a higher weight to larger errors due to the squaring of each term. This makes it more sensitive to outliers or large deviations from the true values.
    \item \textbf{MAE} provides a straightforward arithmetic mean of absolute errors, thus not disproportionately penalizing larger errors compared to smaller ones.
\end{itemize}
When MSE and MAE remain stable across different training data sizes, it suggests that the overall magnitude of errors does not significantly change as more data is used for training. This could imply that adding more training data under these conditions does not necessarily improve the model's ability to predict more accurately in terms of error reduction. It might suggest that the model has reached a plateau in learning from the additional data where the average error remains consistent.

\item \textbf{Pearson Correlation}, on the other hand, measures the strength and direction of a linear relationship between the predicted and actual values. An increase in the Pearson correlation as training data size increases could indicate several things:
\begin{itemize}
    \item As more data is available for training, the model may be getting better at capturing underlying patterns that influence both the scale and trend of the predictions relative to actual outcomes. This doesn’t necessarily mean that the model is becoming more precise in a point-by-point prediction (as indicated by stable MSE and MAE), but rather that it is improving in aligning the direction and trends of its predictions with the actual values.
    \item A higher Pearson correlation means the model's predictions are better aligned with the actual values' variability, even if the absolute errors (magnitude of errors) aren't improving. This could be critical in applications where understanding the direction of changes is more important than the exact errors.
\end{itemize}
\end{enumerate}
Therefore, the observed pattern—stable MSE and MAE but increasing Pearson correlation—suggests that while the precision of the model in absolute terms does not improve with more data, the model's ability to capture the relative movements or trends in the data improves. This distinction is crucial in scenarios where the relationship dynamics between predicted and actual values are more significant than the sheer accuracy of point predictions. It highlights the model’s growing capacity to reflect the true data structure in its predictions over increasing the size of the dataset.
\subsection{Datasets properties}
Dataset characteristics play a pivotal role in model performance. The OssBuilds dataset, with its diversity stemming from four distinct projects, ostensibly presents a more challenging environment for models due to the variability in code patterns and AST structures. However, models generally perform better on this dataset compared to Hadoop, which, despite its larger size, consists of samples from a single project. This counterintuitive result may be attributed to the complexity of Hadoop's ASTs, particularly their depth and diameter, which could pose difficulties for models, especially GNNs, in effectively capturing and propagating information across the tree structure.

\subsection{Transferability across different datasets}
The cross-dataset transferability results highlight the challenges inherent in generalizing models trained on one source code dataset to another. Most models exhibited a decrease in performance when applied to an unfamiliar dataset, underscoring the specificity of learned patterns to the training data's structure and semantics. However, the TreeCNN model and our Dual-Transformer showcased notable resilience, with the latter demonstrating a promising balance between error metrics and correlation, especially on the Hadoop dataset. This suggests that models with sophisticated attention mechanisms, like the Dual-Transformer, may possess an inherent advantage in capturing more generalizable features of source code, transcending dataset-specific idiosyncrasies.

\subsection{models' efficiency across incremental training data sizes}
The exploration into model efficiency with varying sizes of training data revealed an expected trend: model performance generally improved as the amount of training data increased. This trend underscores the importance of data volume in model training, particularly for complex models like GNNs and Transformers, which require substantial data to effectively learn and generalize from the intricate structures of ASTs. The consistent performance improvement of our Dual-Transformer model across incremental training sizes further emphasizes its robustness and the effectiveness of its architectural design in leveraging larger datasets for enhanced source code analysis.
\subsection{Conclusion }
In summary, our findings illuminate the multifaceted nature of source code analysis using machine learning models. The interplay between model architecture (especially the use of attention mechanisms), the level of information granularity, and dataset characteristics significantly influences performance. These insights not only contribute to our understanding of the strengths and limitations of various approaches but also pave the way for future research aimed at optimizing model design and data preprocessing techniques for enhanced source code analysis.

\section{Conclusion}

In this study, we provided an analytical framework to examine the performance of tree-based neural network models in regression tasks. At the heart of our investigation was the introduction of an innovative model predicated on a dual-transformer architecture. This model was meticulously evaluated against an array of models grounded in Graph Neural Network (GNN) and Tree-Based Neural Network (TBNN) paradigms. The rigour of our experimental methodology, applied to two distinct real-world datasets, firmly establishes the dual-transformer model as a superior contender, outshining its counterparts across various error metrics and Pearson correlation indices.

A recurrent theme observed across the evaluated models was the prevalent incorporation of attention mechanisms and a node-level analytical approach within tree structures. This observation accentuates the pivotal role of attention in effectively navigating the structural intricacies inherent in our case study tree.

The contributions of this paper are twofold. Firstly, it introduces a potent model that redefines the benchmark for regression tasks within the realm of source code analysis. Secondly, it facilitates a nuanced comparative analysis of tree-based neural network models, thereby bolstering the understanding of their efficacy and broadening their applicability in practical settings.

Furthermore, the research underscores the Dual-Transformer model's prowess in accurately forecasting source code execution times. By leveraging a dual encoder framework that intricately captures the nuances of source code tokens and AST nodes, our model demonstrates a marked improvement over conventional tree-based neural network approaches. This finding signifies the untapped potential of advanced deep learning architectures in the field of source code analysis, setting a promising direction for future inquiries.

\section*{Acknowledgments}
Anonymized Acknowledgement

\bibliographystyle{IEEEtran}
\bibliography{references}

\end{document}